\documentclass[sigconf,authorversion,nonacm]{acmart}

\usepackage{threeparttable}
\usepackage{balance}

\AtBeginDocument{%
  \providecommand\BibTeX{{%
    \normalfont B\kern-0.5em{\scshape i\kern-0.25em b}\kern-0.8em\TeX}}}

\copyrightyear{2022}
\acmYear{2022}
\setcopyright{acmcopyright}\acmConference[CIKM '22]{Proceedings of the 31st ACM International Conference on Information and Knowledge Management}{October 17--21, 2022}{Atlanta, GA, USA}
\acmBooktitle{Proceedings of the 31st ACM International Conference on Information and Knowledge Management (CIKM '22), October 17--21, 2022, Atlanta, GA, USA}
\acmPrice{15.00}
\acmDOI{10.1145/3511808.3557655}
\acmISBN{978-1-4503-9236-5/22/10}

\begin{document}

\title{Multiple Instance Learning for Uplift Modeling}

\author{Yao Zhao}
\email{nanxiao.zy@antgroup.com}
\affiliation{%
  \institution{Ant Group}
  \city{Hangzhou}
  \country{China}
  \postcode{330100}
}

\author{Haipeng Zhang}
\email{zhanghp@shanghaitech.edu.cn}
\authornote{Corresponding author}
\affiliation{%
  \institution{ShanghaiTech University}
  \city{Shanghai}
  \country{China}
  \postcode{201210}
}

\author{Shiwei Lyu}
\email{lvshiwei.lsw@antgroup.com}
\affiliation{%
  \institution{Ant Group}
  \city{Hangzhou}
  \country{China}
  \postcode{330100}
}

\author{Ruiying Jiang}
\email{ruiying.jry@antgroup.com}
\affiliation{%
  \institution{Ant Group}
  \city{Hangzhou}
  \country{China}
  \postcode{330100}
}

\author{Jinjie Gu}
\email{jinjie.gujj@antgroup.com}
\affiliation{%
  \institution{Ant Group}
  \city{Hangzhou}
  \country{China}
  \postcode{330100}
}

\author{Guannan Zhang}
\email{zgn138592@antgroup.com}
\affiliation{%
  \institution{Ant Group}
  \city{Hangzhou}
  \country{China}
  \postcode{330100}
}

\renewcommand{\shortauthors}{Yao Zhao et al.}

\begin{abstract}

Uplift modeling is widely used in performance marketing to estimate effects of promotion campaigns (e.g., increase of customer retention rate). Since it is impossible to observe outcomes of a recipient in treatment (e.g., receiving a certain promotion) and control (e.g., without promotion) groups simultaneously (i.e., counter-factual), uplift models are mainly trained on instances of treatment and control groups separately to form two models respectively, and uplifts are predicted by the difference of predictions from these two models (i.e., two-model method). When responses are noisy and the treatment effect is fractional, induced individual uplift predictions will be inaccurate, resulting in targeting undesirable customers. Though it is impossible to obtain the ideal ground-truth individual uplifts, known as Individual Treatment Effects (ITEs), alternatively, an average uplift of a group of users, called Average Treatment Effect (ATE), can be observed from experimental deliveries. Upon this, similar to Multiple Instance Learning (MIL) in which each training sample is a bag of instances, our framework sums up individual user uplift predictions for each bag of users as its bag-wise ATE prediction, and regularizes it to its ATE label, thus learning more accurate individual uplifts. Additionally, to amplify the fractional treatment effect, bags are composed of instances with adjacent individual uplift predictions, instead of random instances. Experiments conducted on two datasets show the effectiveness and universality of the proposed framework.

\end{abstract}





\maketitle

\section{Introduction}

Online marketing techniques (e.g., message/in-app banner/push notifications) are major performance marketing tools for service providers \cite{diemert2018large}. However, they sometimes have side effects -- for instance, users bothered by frequent messages may permanently disable app notifications or even uninstall the app. Compared to the brutal force strategy of notifying every potential customer, an optimal strategy would balance the potential gain and the risk of losing a customer’s attention.

Finding such an optimal strategy has been an important topic in marketing research for decades \cite{hansotia2002incremental,truelift}. Intuitively, to decide target customers for a certain marketing campaign, service providers may build a response model (e.g., conversion model) with samples from experimental deliveries, and only touch customers with appreciable response effects \cite{pechyony}. While response modeling seems straightforward, it does not distinguish where the user's response comes from -- if a user visits the app, is it because she is interested in this particular push message, or is it because she is a regular daily visitor anyway? If it is the latter, this push message is unnecessary for her and does not contribute to the number of Daily Active Users (DAUs) for the app. To attribute user responses, uplift modeling \cite{
radcliffe1999differential}, which estimates the difference (i.e., uplift) of response between the treatment group and the control group, is proposed to directly identify customers for whom the treatment is effective.

Compared with Click Through Rate (CTR) prediction tasks that learn from implicit feedback (e.g., expression and click behaviors from online logs instead of explicit rating), uplift modeling faces two challenges: counter-factual nature \cite{gutierrez2017causal} and fractional treatment effect \cite{radcliffe2011real}. For a customer, we can only observe what she does after receiving a push message while we cannot observe what happens if she does not receive it at that time, and vice versa. In other words, one customer can either be in the treatment group or the control group, but never both. Therefore, uplift is the difference between factual and counter-factual labels and cannot be observed directly, making out-of-the-box supervised methods inapplicable. To tackle the problem, several uplift modeling methods use counter-factual learning to obtain estimated counter-factual labels \cite{yoon2018ganite}. However, when estimated labels by generative adversarial networks are not perfectly correct, errors are introduced to subsequent supervised learning steps.

We denote the average uplift of a group of customers as ATE \cite{hatt2021} and the customer-level uplift as ITE \cite{2020Treatment}. Though ground-truth ITEs are impossible to obtain, a ground-truth ATE can be observed from the feedback of an experimental delivery. However, the ATE of an experimental delivery only offers campaign-wise information which helps little for instance-wise learning. To find a balance, we organize instances into 'bags' like Multiple Instance Learning (MIL) suggests. In this way, we can compute bag-wise ATEs and use them to help optimize the overall performance.
The ATE label of a bag is calculated by the difference between the average response probability of treatment and control groups if it contains instances from both groups, and the ATE prediction is the summation of individual uplift predictions from a base model. The bag-wise ATE label can be used to regularize bag-wise ATE prediction, and contribute to learning more accurate individual uplifts. 

Besides the counter-factual problem, the fractional treatment effect problem cannot be ignored. Uplifts are usually much less significant than the main response effect. As shown in Table \ref{tbl.dataset}, for the classic uplift dataset Lenta\footnote{https://www.kaggle.com/mrmorj/bigtarget}, the conversion rate (CVR) is 11.912\% in treatment while 10.257\% in control, meaning that the ATE is fractional (0.755\% vs 10.257\%). Therefore, uplift can be easily shaded by noises of the main response effect \cite{radcliffe2011real}. To overcome the problem, we amplify the bag-wise uplift predictions and labels by aggregating instances into bags according to their uplift predictions. In this way, similar instances are put together to increase the homogeneity within bags and the heterogeneity between bags.

To sum up, our contributions include:
\begin{itemize}
    \item \textbf{Problem.} We identify the problem of counter-factual nature and fractional treatment effect in uplift modeling.
    \item \textbf{Method.} We propose a MIL-enhanced framework to accommodate two-model uplift methods. It uses a bag-wise loss to overcome the counter-factual problem and generates bags by clustering to overcome the fractional treatment effect problem.
    \item \textbf{Evaluation.} Experiments on two datasets suggest consistent improvements over existing SOTA methods.
\end{itemize}

\section{RELATED WORK}

\subsection{Uplift modeling}
The objective of uplift modeling is to estimate Individual Treatment Effect (ITE), given the individual’s features $X$,
\begin{equation}
h(x)=E[Y=1|X=x, T=1]-E[Y=1|X=x, T=0]
\end{equation}
The optimal strategy is simply choosing customers with higher ITE scores. The biggest challenge is that the two outcomes are mutually exclusive for an individual. The counter-factual nature makes traditional supervised learning inapplicable here.

Tackling this problem, algorithms can be roughly classified into three categories: two-model approach, class transformation approach and modeling uplift directly \cite{gutierrez2017causal}. Two-model methods, as a mainstream approach with better performance \cite{radcliffe1999differential, zaniewicz2013support, betlei2021uplift, dragonnet}, construct a model $E(Y = 1|X,T = 1)$ using only the treatment data and another $E(Y = 1|X,T = 0)$ with only the control data, and calculate uplift predictions as their difference. Its advantage is obvious: any supervised learning method fits in to model the uplift, especially the neural network methods. However, the two models are optimized separately, without knowing each other. To overcome this, \cite{betlei2018uplift} proposes two methods: dependent data representation and shared data representation. To link two models, the former method predicts treatment responses by features of users and predictions of the control group model while the latter lets two models share a part of output. 
To tackle the counter-factual nature of uplift modeling, \cite{shalit2017estimating, yoon2018ganite,cevea} take counter-factual learning frameworks to estimate the response of treatment distribution, while \cite{yamane2018uplift} proposes a direct individual uplift estimating method for the situation that one of treatment flag and outcome is inaccessible. Class transformation methods \cite{jaskowski2012uplift,athey2015machine} focus on transforming counter-factual uplift labels to factual labels and then apply supervised learning. \cite{jaskowski2012uplift} reverses the labels in the control group and keeps labels in the treatment group unchanged to induce a transformed label $Z$, and predicts ITE by $2E(Z=1|X=x)-1$. \cite{athey2015machine} proposes a generalized method considering unbalanced treatment assignment. Transformation methods are usually not unbiased estimations, which are not convenient for estimating outcomes. Modeling the uplift directly \cite{chickering2013decision, guelman2012random, hansotia2002incremental} is another path. Uplift can be easily captured by decision-tree-based methods so many algorithms in this category are tree-based. In \cite{hansotia2002incremental}, the authors maximize the desired quantity directly. The proposed algorithm selects tests maximizing the difference between the differences between treatment and control success probabilities in the left and right subtrees. 
To archive better performance with tree-based methods, customized split functions \cite{rzepakowski2010decision} and ensemble methods are proposed \cite{guelman2012random, zhao2017practically, soltys2015ensemble}. 
However, the effectiveness of tree-based methods is limited by the greedy splitting criterion, which is sub-optimal for most cases, and their scalability is not comparable with neural-network-based methods.

\subsection{MIL}

In MIL \cite{dietterich1997solving}, a bag is a collection of instances, and it is positively labeled if it contains at least one positive instance. Otherwise, it is labeled as a negative bag. A training set comprises labeled bags but without labeled instances, and a classic task is to predict the labels of unseen bags or unlabeled instances. MIL is introduced for drug activity prediction problem \cite{dietterich1997solving}, and then widely applied in the bioinformatics \cite{bandyopadhyay2015mbstar}, computer vision \cite{shamsolmoali2020amil}, text classification \cite{chai2017,pappas2014explaining}, of which weak or holistic labels usually are accessible while instance-level labels are absent. MIL methods can be grouped into two categories: bag-space and instance-space. Bag-space methods \cite{feng2021multiple, ma2021multi,wu2014} represent a bag by meta-information or aggregated detailed features, and some define distance within bags directly. Pooling \cite{feng2017deep, zhu2017deep, ilse2018attention} and attention\cite{ma2021multi} are popular frameworks to capture instance and bag similarity. Instance-space methods \cite{zhang2001dd, zhu2017deep,luo2020weakly} predict bag labels by aggregating instance labels, which estimates instance-wise labels and ignores the interaction of instances in a bag. Expectation-maximization framework \cite{zhang2001dd,luo2020weakly} is a widely used method to identify instance labels. In contrast to bag classification, several studies \cite{liu2012key, kotzias2015group, peng2019address} focus on instance classification with known bag labels. Though sharing some similarities with our scenario, they try to solve binary classification problems with binary bag labels, which are essentially different compared to our situation. Uplifts are in fact continuous and the initial coarse uplift predictions are provided by a base model, making these methods inapplicable directly.

\section{MIL UPLIFT MODELING}

\subsection{Methodology}

Our proposed MIL framework is applied to two-model uplift methods, which enjoy higher popularity and archive better performance recently \cite{betlei2018uplift,shalit2017estimating,yoon2018ganite}. We denote a two-model method as a base model. The framework cooperates with a base model to overcome counter-factual nature and fractional treatment effect problems, in three steps: 1) Train a base model with an arbitrary two-model method. 2) Use the base model to predict uplifts of instances in a mini-batch, and cluster instances with adjacent uplifts into bags. Specifically, we cluster instances in a mini-batch by their uplifts and pack $j$ adjacent instances (without overlaps) into an equal-sized bag. 3) Calculate bag-wise ATE labels and predictions, and add a distance loss function to the loss functions of the base model for training. We will introduce relevant calculations in detail.

\subsubsection{Bag-Wise ATE Label}

Conventionally, an ATE label is calculated as the difference between the average response probability of treatment and control instances. However, it may cause unstableness in optimization when the bag size is not large enough or the two groups are highly imbalanced. To alleviate the problem, we calculate a bag-wise expected summed response of treatment and control instead of the average response:
\begin{equation}
\begin{split}
y_{bag} = \Sigma_{i\in T} \frac{y^t_i}{u_t} - \Sigma_{j\in C} \frac{y^c_j}{1-u_t}
\end{split}
\end{equation}
where $y^t_{i}$ and $y^c_{j}$ denote response label of the $i$-th user in treatment group and the $j$-th user in control group in a bag, respectively. $u_t=\#treatment/(\#treatment+\#control)$ is the treatment ratio in a mini-batch.

\subsubsection{Bag-Wise ATE Prediction}

Intuitively, bag-wise ATE prediction can be calculated by the summation of instance-wise uplift predictions in a bag: $h_{bag} = \Sigma_i (p^t_{i} - p^c_{i})$, where $p^t_{i}$ and $p^c_{i}$ denote the response probabilities of treatment group and control group, respectively, for the $i$-th user in the bag. Since bag-wise ATE labels are calculated by weighted summation, we decide to calculate the bag-wise ATE predictions by weighted summation as well, for better empirical performance:
\begin{equation}
\begin{split}
h_{bag} = \Sigma_{i\in \mathcal{T}} \frac{p^t_{i}}{u_t} -  \Sigma_{j\in \mathcal{C}} \frac{p^c_{j}}{1-u_t}
\end{split}
\end{equation}

\subsubsection{Loss Function}
Our optimization object of the additional loss function is to minimize the difference between the bag-wise ATE labels and ATE predictions:
\begin{equation}
\begin{split}
L_{mil} & =\Sigma_{k} (y^k_{bag} - h^k_{bag} )^2 \\
L & = L_{base} + \alpha L_{mil} \\
\end{split}
\end{equation}
where $L_{base}$ is a loss function of the base model. $y^k_{bag}$ and $h^k_{bag}$ are the ATE label and the ATE prediction of $k$-th bag, respectively. 

\subsubsection{Algorithm Analysis}

For a two-model based model, two assumptions are preset: both outputs of the two models are unbiased estimations as outcomes, and the noise $\epsilon$ follows i.i.d normal distribution with variance $\sigma^2$:
\begin{equation}
\begin{split}
& E[p(Y=1|X=x,T=t)] = E[Y|X=x,T=t] \\
& p(Y=1|X=x,T=t) = E[p(Y=1|X=x,T=t)] + \epsilon \\
& \quad \quad \quad s.t.  \quad \epsilon \sim N(0, \sigma) \\
\end{split}
\end{equation}
With the above assumptions, we can easily find out that the individual uplift prediction is also an unbiased estimation, with noise variance $2\sigma^2$. Therefore, we only need to reduce the estimation variance to obtain better performance, which can be archived by applying additional MIL loss, since they are equivalent:
\begin{equation}
\label{eq.equality}
\begin{split}
L_{mil} & = \Sigma_i (h^i_{bag} - y^i_{bag} )^2 = \Sigma_i ( \Sigma_{j} (y_{ij} + \epsilon_{ij})  - \Sigma_{j} y_{ij} )^2 \\ 
& = \Sigma_{ij} \epsilon_{ij}^2 = \sigma^2
\end{split}
\end{equation}
Its effectiveness is further checked in Section \ref{sec.exp}. 


\section{EXPERIMENTAL EVALUATION}

\subsection{Datasets}

We conduct experiments on two public real-life datasets. CRITEO-uplift v2 \cite{diemert2018large} (denoted as CRITEO) is a large-scale dataset from an Internet advertisement scenario. Lenta is another dataset collected from a retail scenario, where the treatment group is informed with SMS and the response is visiting. The two datasets are summarized in Table \ref{tbl.dataset}. 
\begin{table}[ht]
\caption{Summary of datasets.}
\label{tbl.dataset}
\begin{tabular}{ccc}
	\toprule  
	Dataset & Lenta & CRITEO \\
	\midrule  
	Size & 687,029 & 13,979,592 \\
	Features & 194 & 12 \\
	Group $\mathcal{T}$ ratio & 0.75 & 0.85 \\
	Positive class ratio & 0.108 & 0.047 \\
	Pos. class ratio in group $\mathcal{T}$ & 0.11012 & 0.04854 \\
	Pos. class ratio in group $\mathcal{C}$ & 0.10257 & 0.03820 \\
	Average uplift & 0.00755 & 0.01034 \\
	\bottomrule 
\end{tabular}
\end{table}

\subsection{Evaluation metrics}

For a given customer, we know only one outcome, with or without the treatment. Therefore, the evaluation task is more challenging than traditional machine learning. One practical evaluation approach is to visualize model performance using uplift curves~\cite{soltys2015ensemble}. As with Receiver Operating Characteristic (ROC) curves in classification, we use the Area Under Uplift Curve (AUUC) to summarize uplift model performance with a single metric. AUUC varies whether treated and control groups are ranked separately or jointly \cite{devriendt2020learning}, we use the separate one for robustness on imbalanced treatment and control conditions. 

\subsection{Implementation details}

The hidden layers are (1024, 512, 256) with ReLU activation function and the optimizer is Adam with $\beta_1$=0.9 and $\beta_2$=0.999, early stopping is applied to prevent over-fitting (without extra notes, parameters are applied in all methods and both datasets). The learning rate is 0.001 for the Lenta dataset and 0.0001 for CRITEO. The methods DDR \cite{betlei2018uplift} +MIL and SDR \cite{betlei2018uplift} +MIL with Lenta dataset use 0.01 for MIL loss weight while others use 0.001. Training steps are 10,000 and 40,000 for Lenta and Criteo datasets respectively. In GANITE \cite{yoon2018ganite}, a generator and a discriminator are trained in turn with 50 steps in the first 5,000 and 20,000 steps for Lenta and Criteo datasets respectively. The batch size is 1,024 and the bag size is 64. 
To avoid the effect caused by the random initial weight in a neural network, we repeat every experiment at least 5 times, and report the average performance. 

There are two common practices to generate two outputs for control and treatment groups: one is by a model with two output nodes and the other is by two separated models. 
We believe the former would have better performance attributed to thorough training, since hidden layers in a two-output-node network can be trained with all instances while hidden layers in separate towers are only trained with corresponding instances. With this output node architecture and the above hyper-parameters, the baseline models implemented in our experiments generally perform better than their original versions. For example, AUUC of SDR with the CRITEO dataset is 0.00923 in the original paper \cite{betlei2021uplift}, while being much higher (0.00976) with our implementation.

\subsection{Experiment results}\label{sec.exp}
We conduct experiments regarding state-of-the-art methods w/o MIL-boosting. As shown in Table \ref{tbl.model}, our method outperforms all base models on both datasets. On the Lenta dataset, the best AUUC of our method surpasses that of the baselines by a large margin (0.007839 for GANITE+MIL vs 0.006763 for TARNet). Meanwhile, when we examine the enhancement brought by MIL w.r.t. a given baseline, we see an improvement as much as 25\% (GANITE+MIL vs GANITE in Lenta). In the CRITEO dataset, the best AUUC of our method (0.009895) and that of the baselines (0.009849) seem comparable. However, the best baseline's AUUC (0.009849) is already very near the ATE of this dataset (0.01034), which can be viewed as an approximate upper bound of AUUC. Taking this into consideration, our method's improvement is non-negligible. Among all comparisons, the largest improvements are found for the relatively inferior base models which intuitively have larger margins for enhancements. Additionally, we notice that compared with base models, MIL-boosting methods have smaller variances in general, except GANITE in Lenta and TM in CRITEO. As mentioned above, the reduced variances might be the result of introducing additional MIL loss in our design.

\begin{table}[ht]
\caption{AUUC of different models w/o MIL loss.}
\small
\label{tbl.model}
\begin{tabular}{cccc}
	\toprule  
Dataset & Model & Original ($\times 0.001$) & +MIL ($\times 0.001$)\\ 
	\midrule  
Lenta & TM & { 5.694±0.631 } & 6.376±0.301 \\
Lenta & DDR \cite{betlei2018uplift}  & 5.922±0.572 & 6.072±0.252 \\
Lenta & SDR \cite{betlei2018uplift} & 5.952±0.412 & 6.442±0.250 \\
Lenta & TARNet \cite{shalit2017estimating}  & 6.763±0.410 & 7.111±0.193 \\
Lenta & CFR \cite{shalit2017estimating}  & 5.808±0.435 & 6.757±0.075 \\
Lenta & GANITE \cite{yoon2018ganite}  & 6.260±0.077 & 7.839±0.113 \\
	\midrule  
CRITEO & TM & 9.678±0.113 & 9.827±0.186 \\
CRITEO & DDR \cite{betlei2018uplift}  & 9.491±0.132 & 9.800±0.019 \\
CRITEO & SDR \cite{betlei2018uplift} & 9.755±0.133 & 9.880±0.079 \\
CRITEO & TARNet \cite{shalit2017estimating}  & 9.833±0.096 & 9.895±0.003 \\
CRITEO & CFR \cite{shalit2017estimating}  & 9.849±0.077 & 9.853±0.063 \\
CRITEO & GANITE \cite{yoon2018ganite}  & 9.473±0.181 & 9.742±0.018 \\

\bottomrule 
\end{tabular}
\end{table}

The performance of different hyper-parameters including bag size and MIL loss weight, is listed in Table \ref{tbl.parameters}. Under limited resources, we only search parameters with TARNet model and Lenta dataset by running 3 times (except the best parameter set with 5 runs). The parameter set with batch size=1,024, bag size=64, MIL loss weight=0.001, performs best and is therefore applied to all MIL-improved methods. We can see that MIL-loss is sensitive to bag size and loss weight. Large MIL loss weight decays the performance, especially with a small bag size. A possible explanation is that a small bag size introduces large variance (noise) of estimated uplift labels and a large loss weight will enhance the impact.

\begin{table}[ht]
\setlength\tabcolsep{3pt}
\caption{AUUC of different bag-size and MIL-loss weight with TARNet model and Lenta dataset.}
\small
\label{tbl.parameters}
\begin{threeparttable}[b]
\begin{tabular}{lccc}
	\toprule  
	Sizes\tnote{1} & $\alpha=0.0001 (\times 0.001)$ & $\alpha=0.001 (\times 0.001)$ & $\alpha=0.01 (\times 0.001)$ \\ 
	\midrule  
    1,024*8 & 6.53±0.39 & 6.43±0.51 & 5.99±0.51 \\
    1,024*16 & 6.30±0.38 & 6.41±0.38 & 5.65±0.03 \\
    1,024*32 & 6.36±0.38 & 6.44±0.26 & 5.79±0.18 \\
    1,024*64 & 6.01±0.31 & 7.11±0.19 & 6.44±0.29 \\
    1,024*128 & 6.28±0.55 & 6.87±0.40 & 6.38±0.40 \\
	\bottomrule 
\end{tabular}
\begin{tablenotes}
\item[1] Sizes denotes $batch\_size\times bag\_size$
\end{tablenotes}
\end{threeparttable}
\end{table}

To verify the necessity of a base model and clustering trick, we conduct further experiments with the CRITEO dataset (see Table \ref{tbl.sort}). Without the clustering trick, the performance decays considerably for both datasets. If we only use the MIL loss for training, the results show significant drops. These indicate these design considerations are actually effective in our framework.

\begin{table}[ht]
	\caption{AUUC of models without clustering trick or a base model loss function. The base model is TARNet.}
    \small
	\label{tbl.sort}
\begin{tabular}{ccc}
	\toprule  
	Model & Lenta ($\times 0.001$) & CRITEO ($\times 0.001$) \\ 
	\midrule  
	w/o base model & 5.284±0.618 & 8.895±0.787  \\
	w/o clustering & 6.059±0.697 & 9.853±0.017 \\
	Proposed & \textbf{7.111±0.193} & \textbf{9.895±0.003} \\
	\bottomrule 
\end{tabular}
\end{table}

\section{CONCLUSION}

We design an effective MIL-enhanced framework to accommodate two-model uplift models, by introducing a new bag-wise regularizing loss function. As an easy add-on, the framework overcomes counter-factual nature and fractional treatment effect problems in uplift modeling and brings considerable improvements to mainstream models. For future work, we plan to determine the hyper-parameters, including the learnable loss weight and gradient-sensitive bag size, in a more efficient and automatic fashion.

\clearpage

\bibliographystyle{ACM-Reference-Format}
\balance
\bibliography{uplift}

\clearpage

\appendix

\section{Algorithm Analysis}

For a two-model based model, two assumptions are preset: both outputs of the two models are unbiased estimations as outcomes, and the noise $\epsilon$ follows i.i.d normal distribution with variance $\sigma^2$:
\begin{equation}
\begin{split}
& E[p(Y=1|X=x,T=t)] = E[Y|X=x,T=t] \\
& p(Y=1|X=x,T=t) = E[p(Y=1|X=x,T=t)] + \epsilon \\
& \quad \quad \quad s.t.  \quad \epsilon \sim N(0, \sigma) \\
\end{split}
\end{equation}
With the above assumptions, we can easily find out that the individual uplift prediction is also an unbiased estimation, with noise variance $2\sigma^2$. Therefore, we only need to reduce the estimation variance to obtain better performance, which can be archived by applying additional MIL loss, since they are equivalent:
\begin{equation}
\label{eq.analysis}
\begin{split}
L_{mil} & = \Sigma_i (h^i_{bag} - y^i_{bag} )^2 = \Sigma_i ( \Sigma_{j} (y_{ij} + \epsilon_{ij})  - \Sigma_{j} y_{ij} )^2 \\ 
& = \Sigma_{ij} \epsilon_{ij}^2 = \sigma^2
\end{split}
\end{equation}
Its effectiveness is further checked in Section \ref{sec.exp}. 

In the Eq.\ref{eq.analysis}, we assume implicitly that the bag-wise ATE labels are accurate. In fact, they may also contain errors as they are calculated by expectation. However, the conclusion also holds since the errors are zero mean variables and can be viewed as noises in SGD. The convergence of SGD is guaranteed by the central limit theorem if an optimizer without noises can converge \cite{chen2016statistical,anastasiou2019normal}.

\section{AUUC curve}

\begin{figure}[ht]
	\centering
	\includegraphics[width=3.5in]{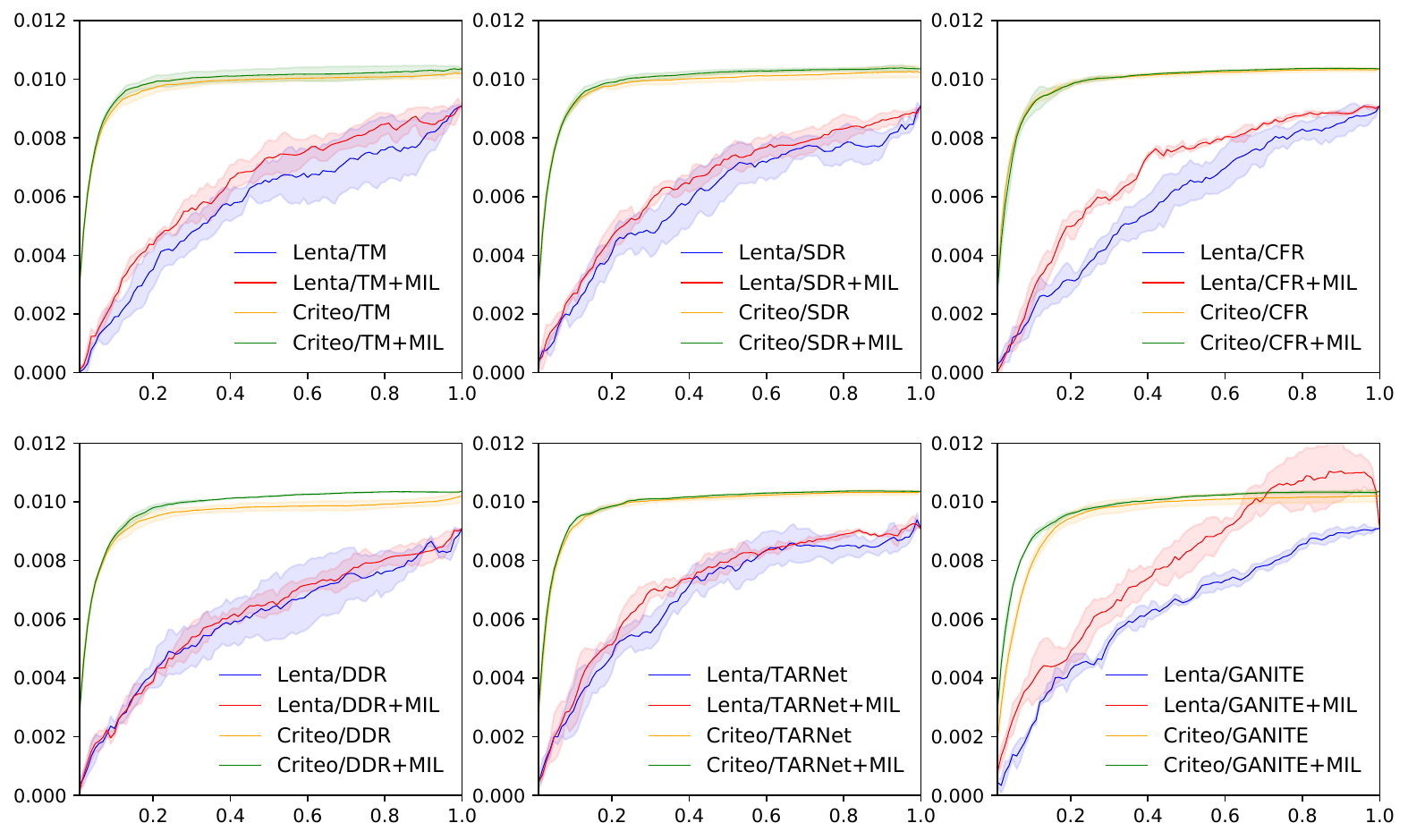}
	\caption{Uplift curves of models w/o MIL loss.}
	\label{fig.curve}
\end{figure}

AUUC curves of comparison models are demonstrated in Figure \ref{fig.curve}. In the figure, higher AUUC curves suggest that with different proportions of users, our method always has a better performance. Especially in usual real-world scenarios when we target a small proportion of users, our method is able to retain superior performance.

\end{document}